\documentclass{Interspeech}

\interspeechcameraready

\title{Prominence-aware automatic speech recognition for conversational speech}

\author[affiliation={1}]{Julian}{Linke}
\author[affiliation={1}]{Barbara}{Schuppler}

\affiliation{Signal Processing and Speech Communication Laboratory}{Graz University of Technology}{Austria}
\email{linke@tugraz.at, b.schuppler@tugraz.at}
\keywords{prominence detection, automatic speech recognition, prominence-aware ASR, wav2vec2}

\usepackage{comment}

\usepackage{multirow}
\usepackage{subcaption}
\usepackage{lipsum}
\usepackage[export]{adjustbox}

\begin{document}

\maketitle

\begin{abstract}
This paper investigates prominence-aware automatic speech recognition (ASR) by combining prominence detection and speech recognition for conversational Austrian German. First, prominence detectors were developed by fine-tuning wav2vec2 models to classify word-level prominence. The detector was then used to automatically annotate prosodic prominence in a large corpus. Based on those annotations, we trained novel prominence-aware ASR systems that simultaneously transcribe words and their prominence levels. The integration of prominence information did not change performance compared to our baseline ASR system, while reaching a prominence detection accuracy of 85.53\% for utterances where the recognized word sequence was correct. This paper shows that transformer-based models can effectively encode prosodic information and represents a novel contribution to prosody-enhanced ASR, with potential applications for linguistic research and prosody-informed dialogue systems.
\end{abstract}

\section{Introduction}

Prosodic prominence is a complex phenomenon that manifests through multiple acoustic and perceptual dimensions \cite{Wagner2005}. Accurate prosodic prominence detection is crucial for applications like speech synthesis, language learning tools, and clinical voice analysis, yet
remains elusive in spontaneous speech contexts. Various approaches have emerged to study syllable- or word-level prosodic prominence for the development of automatic prosodic annotation tools that map acoustic, lexical and syntactic features to prominence \cite{ANALOR, Narayanan2008, Christodoulides2017, linke2023wordlevel}. What these systems have in common is that they require some sort of annotation prior to the prominence annotation, for instance at the level of phone, syllable or word segmentations. Based on these annotations, acoustic prosodic features are then extracted (e.g., F0-, RMS and duration-related) and subsequently fed into different types of classifiers (e.g., Random Forests in \cite{linke2023wordlevel}). While these approaches yield  classification performances in the range of human-inter-rater agreements, for more complex spontaneous and conversational speech, such tools encounter their limits, especially with respect to the following two critical challenges that can cause data loss:
\par
First, in spontaneous speech voice quality tends to vary much more than in read speech, with frequent occurrences of breathy and creaky voice, which may function in the dialogue for signaling a turn-hold, or to convey other paraliguistic or pragmatic meaning to the conversation. Furthermore, overlapping speech occurs frequently (e.g., in approx. $42\%$ of all GRASS utterances \cite{schuppler2017corpus}). These spontaneous speech characteristics have an effect on acoustic feature extraction and results in unreliable extraction of features derived from the extracted RMS and F0 contours (i.e., failures in detecting peaks and/or valleys in short segments). Another limitation of traditional approaches is their dependence on accurate phone or syllable segmentations. While for read speech, automatically generated segmentations (i.e., by means of an ASR system in Forced Alingment mode) are comparable in accuracy to manually created phone segmentations, this is not the case for spontaneous speech. The accurate segmentation, however, is important not only for the exact computation of durational features (e.g., local and global articulation rates), but also for the above mentioned extraction of F0- and RMS related features.  
\par
In this paper, we present a novel approach to prominence annotation, that  relies neither on an existing orthographic annotation nor on phone- or 
syllable-level segmentations, nor on the extraction of prosodic features. Instead, prominence annotation is performed simultaneously with the automatic word-level transcription by means of a transformer-based ASR system. Instead of relying on a set of error-prone F0 extractions, our prominence detection tool relies on self-supervised speech representations extracted from raw audio by means of wav2vec2 \cite{baevski2020wav2vec}. Since this  system combines automatic prominence detection with automatic speech recognition, we use  the term \emph{prominence-aware ASR} for this innovative approach.
\par

The wav2vec2 architecture stands as a robust framework for extracting self-supervised speech representations from raw unlabeled audio data, making it a particularly suitable foundation for various kinds of speech processing tasks through its hierarchical encoding of both segmental and suprasegmental features. The convolutional layers capture interpretable phonetic features that align with classical phonetic knowledge, while the transformer layers organize acoustic-phonetic information in other ways that enable excellent phone classification \cite{tenbosch23_interspeech}. With respect to prosody, a recent study found that transformer layers hierarchically integrate syllable-level stress patterns \cite{bentum24_interspeech} and that boundary detection is achievable with an
F1-score of 83\% on within-sentence prosodic boundaries \cite{Kunesova2022DetectionOP}. A comparison of wav2vec2 codebook usage revealed that codebook entries do not only encode languages \cite{conneau21_interspeech}\footnote{A. Conneau, A. Baevski, R. Collobert, A. Mohamed, and M. Auli, “Unsupervised Cross-Lingual Representation Learning for Speech Recognition,” \textit{	arXiv:2006.13979}.} but also language varieties, speaking styles and speakers \cite{linke23_interspeech}. This codebook versatility directly supports our hypothesis that wav2vec2 embeddings inherently encode prosodic information usable for prominence detection. Given that wav2vec2 additionally achieves good WERs also in low-resource conditions (e.g., \cite{baevski2020wav2vec}), we find it a suitable framework for developing our prominence-aware ASR system.
\par

For the development of our prominence-aware ASR system, we have as starting point the a small subset of manually-annotated prominence levels from the GRASS CS corpus, comprising approx. 4.4 hours of annotated speech. Using this data, we first develop a prominence detection tool based on fine-tuned self-supervised representations that can distinguish prominence levels. We use the prominence detector to then annotate the entire GRASS corpus automatically. Thereby we receive the amount of prominence labels needed to train an ASR system that is then able to annotate word sequences along with their prominence levels in an integrated fashion \cite{linkePhD}.

\begin{table}[t!]
\caption{Overview of the used Austrian German speech data for the prominence detectors $\text{PDET}_{02}$ and $\text{PDET}_{012}$. The table shows orthography and corresponding reference examples while the prominence detectors were exclusively trained using the references. }
\centering
\begin{tabular}{c|c|c}
\multicolumn{1}{c|}{\textbf{Type}} & \textbf{Orthography} & \textbf{Reference}  \\
\midrule
$\text{PDET}_{02}$ & \textbar\phantom{0} sie hat \phantom{0}\textbar\phantom{0} erzählt \phantom{0}\textbar & \textbar\phantom{0}0 \textbar\phantom{0}2 \textbar \\
$\text{PDET}_{012}$ & \textbar\phantom{0} wah \phantom{0}\textbar\phantom{0} voll \phantom{0}\textbar\phantom{0} nett \phantom{0}\textbar & \textbar\phantom{0}0 \textbar\phantom{0}2 \textbar\phantom{0}1 \textbar \\
\end{tabular}
\label{tab:PDETOverview}
\end{table}

\section{Prominence Detection}
\label{ASRProminence:PDET}

\subsection{Materials and Methods}
\label{ASRProminence:PDET:M&M}

\subsubsection{GRASS corpus}
\begin{figure}[t]
    \centering
    \begin{subfigure}{0.15\textwidth}        \includegraphics[trim={1.5cm 0cm 0cm 0cm},keepaspectratio,valign=t,width=1\textwidth]{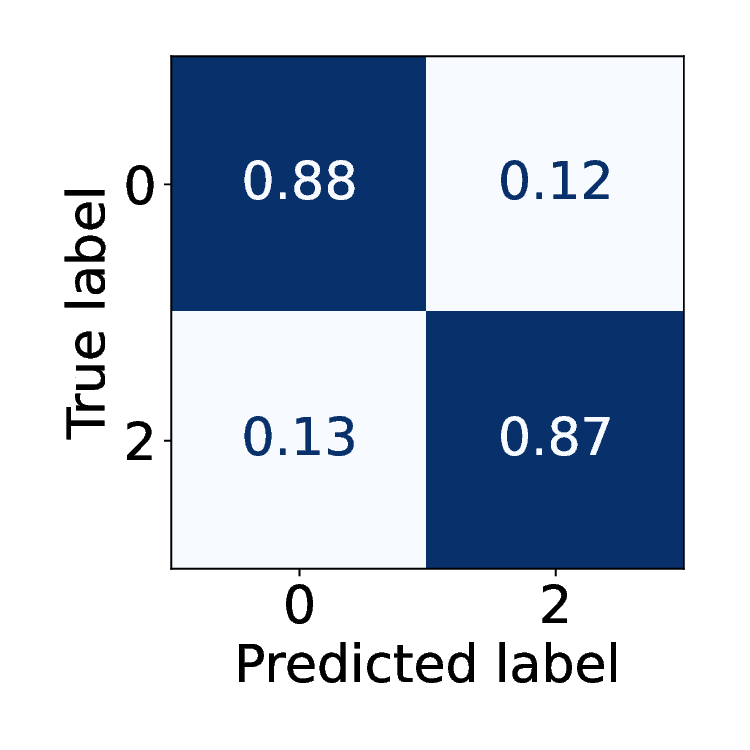}
    \end{subfigure}
    \hfill
    \begin{subfigure}{0.15\textwidth}
\includegraphics[trim={1.5cm 0cm 0cm 0cm},keepaspectratio,valign=t,width=1\textwidth]{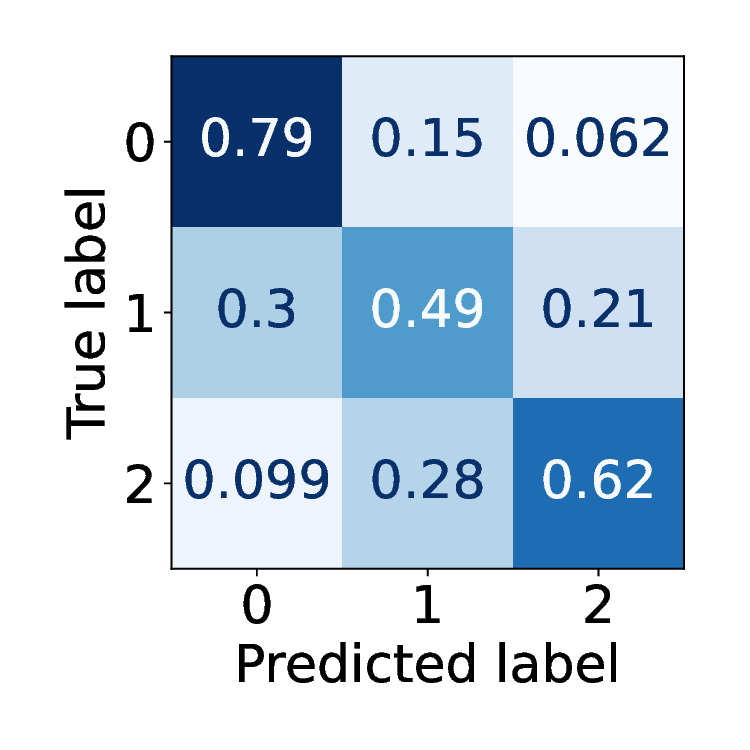}
    \end{subfigure}
    \hfill
    \begin{subfigure}{0.15\textwidth}
    \raisebox{-0.16cm}{\includegraphics[trim={0.5cm 0cm 0cm 0cm},keepaspectratio,,valign=t,width=.94\textwidth]{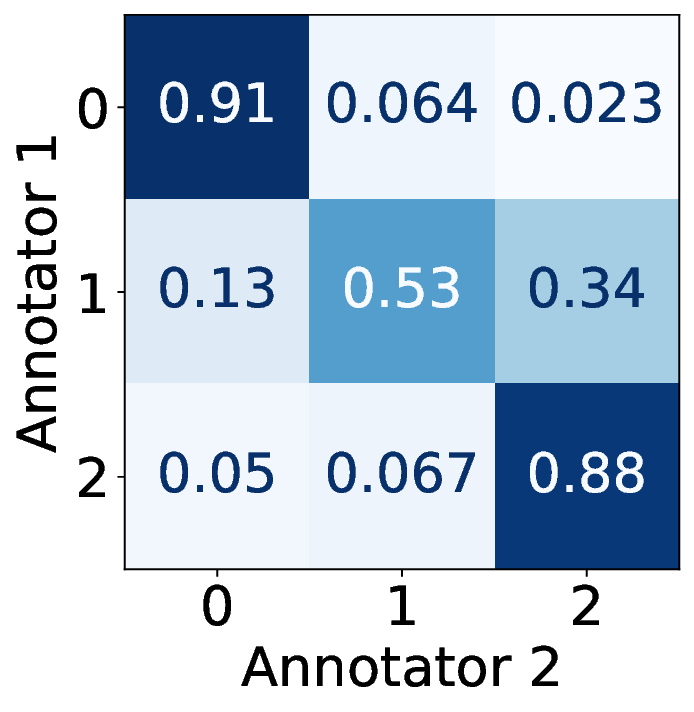}}
    \end{subfigure}
    \caption{Confusion matrices derived from prominence detectors $\text{PDET}_{02}$ (left) and $\text{PDET}_{012}$ (middle) for conversation with ID 004M024F and corresponding confusion matrix of human annotators (right). Results of the prominence detectors refer only to words of utterances where alignment between human-annotated word boundaries and detection-annotated word boundaries was possible.}
    \label{fig:CM_PDET_004M024F}
\end{figure}

Our experiments are based on the \textit{Graz corpus of Read and Spontaneous Speech} (GRASS, \cite{schuppler2014grass, GRASS2017}), featuring 19 face-to-face conversations between two closely acquainted native speakers of Austrian German. The complete corpus was annotated manually with orthographic transcriptions, and a subset was additionally annotated prosodically using the KIM annotation system \cite{KIM}. The prosodic annotations were created by phonetically trained transcribers for a total of $4944$ utterances including $15664$ word tokens from $34$ speakers. The prominence annotations distinguished the prominence levels $0$ (no prominence; \emph{PL0}), $1$ (weak prominence; \emph{PL1}), $2$ (strong prominence) and $3$ (emphatic prominence). Prominence levels $2$ and $3$ were combined as \emph{PL2}. Respective inter-annotator agreements (cf. confusion matrix of human annotators in Fig.~\ref{fig:CM_PDET_004M024F}) were 0.72 (\emph{PL0}/\emph{PL1}), 0.92 (\emph{PL0}/\emph{PL2}) and 0.57 (\emph{PL1}/\emph{PL2}).

\subsubsection{Prominence detection}
\par % prominence detector
Prominence detectors were developed by fine-tuning the wav2vec2 XLSR model \cite{baevski2020wav2vec,conneau21_interspeech} with the prominence-annotated utterances and a CTC loss \cite{Graves2006}. More precisely, we trained two separate prominence detectors $\text{PDET}_{02}$ and $\text{PDET}_{012}$, where the first detector classified two prominence levels (\emph{PL0} vs. \emph{PL2}) and the second detector classified three prominence levels (\emph{PL0} vs. \emph{PL1} vs. \emph{PL2}). Tab.~\ref{tab:PDETOverview} gives an overview of the training data with respect to the two types of models. The reference text for training included only the resulting prominence levels as single numbers plus word boundary markers (''\textbar''). For $\text{PDET}_{02}$ the training data included 1770 utterances with $2.09\pm1.39$ tokens and for $\text{PDET}_{012}$ 4944 utterances with $3.17\pm2.13$ tokens. Note that prominence annotations referred to prosodic words (e.g., the prosodic word ''\textbar\phantom{0}sie hat \textbar" was annotated as \emph{PL0}). For each type of detector, we performed 10-fold cross-validation in order to test the generalization ability of the prominence detectors and provide corresponding accuracy means and standard deviations. Additionally, we trained models for one held-out test conversation (i.e., conversation with ID \text{004M024F}). For evaluation, we compare 1) prominence detection error rates \text{(PER)} calculated similarly as word error rates while considering only prominence levels 
%and word boundary markers 
and 2) accuracies, F1-scores and recalls for prominent words but only if an alignment between human-annotated word boundaries and detection-annotated word boundaries was possible with respect to each utterance. 

In a final step, the entire GRCS component was automatically annotated twice with the final prominence detectors $\text{PDET}_{02}$ and $\text{PDET}_{012}$. For each utterance, if the detection results aligned with the word boundaries of given forced alignments of a Kaldi system \cite{linke2023using} only these words were automatically annotated with a prominence level (i.e., with respect to each speaker approx. $52.06\%\pm8.57\%$ ($\text{PDET}_{012}$) and $42.3\%\pm8.4\%$ ($\text{PDET}_{02}$) of the utterances were aligned). For consistency, the automatic annotation of the entire GRCS component was based on word boundaries coming from forced alignments as human-annotated word boundaries are only available for the smaller prominence-annotated subset. These automatically annotated words were then utilized as additional information for prominence-aware ASR training. 

\begin{table*}[t!]
\caption{Prominence detection results of prominence detectors $\text{PDET}_{02}$ and $\text{PDET}_{012}$ for two test conditions. The prominence error rates (\textbf{PER}) $[\%]$ and accuracies $[\%]$ of 10-fold CV results are shown with mean and standard deviations. The \textbf{PER} was calculated for all utterances of a test split. The ratio of possible alignments given correct word boundaries of an utterance for each test split (\textbf{\%Aligned}) explains for which amount of utterances the word-level accuracy measurements could be calculated (\textbf{Accuracy}).}
\centering
\begin{tabular}{c|c|c|c|c}
  \textbf{Type} &  \textbf{Test set} & \textbf{PER} & \textbf{\%Aligned} & \textbf{Accuracy} \\ \hline
\midrule
\multirow{2}{*}{$\begin{array}{@{}c@{}} \text{PDET}_{012} \end{array}$} & 10-fold CV & $36.54\pm0.92$  &  $66.80\pm1.66$ &  $69.45\pm2.11$ \\
 & 004M024F & $41.02$ & $64.34$ &  $64.97$ \\ \hline
\multirow{2}{*}{$\begin{array}{@{}c@{}} \text{PDET}_{02} \end{array}$} & 10-fold CV &   $24.83\pm1.79$ &   $69.56\pm3.00$ &  $89.72\pm3.26$ \\
 & 004M024F &  $29.58$ & $63.48$ &  $87.40$  \\ 
 \midrule 
 $\text{ASR}_{02}(\text{PDET}_{02})$  & 004M024F &  $65.42$ & $52.17$ &  $85.53$   \\ %\hlin
\end{tabular}
\label{tab:PDETResults}
\end{table*}

\subsection{Results for prominence detection}
Tab.~\ref{tab:PDETResults} shows prominence detection results for all types of models. For $\text{PDET}_{02}$ we achieved \text{PERs} of $24.83\% \pm 1.79\%$ (10-fold  CV) and $29.58\%$ (004M024F). For this model, it was possible to align $69.56\% \pm 3.00\%$ (10-fold  CV) or $63.48\%$ (004M024F) of the utterances with respect to the detected word boundaries. For these words, we achieved accuracies of $89.72\% \pm 3.26\%$ (10-fold  CV) or $87.40\%$ (004M024F). 
\par
In contrast, for $\text{PDET}_{012}$ we achieved worse \text{PERs} of $36.54\% \pm 0.92\%$ (10-fold  CV) and $41.02\%$ (004M024F). This time, it was possible to align $66.80\% \pm 1.66\%$ (10-fold  CV) or $64.34\%$ (004M024F) of the utterances with respect to the detected word boundaries. Furthermore, we achieved worse accuracies of $69.45\% \pm 2.11\%$ (10-fold  CV) or $64.97\%$ (004M024F). 
\par
Confusion matrices in Fig.~\ref{fig:CM_PDET_004M024F} illustrate in more detail results for conversation with ID 004M024F. With respect to recalls of $\text{PDET}_{02}$ (for 119 aligned words out of 73 utterances), it can be seen that $84\%$ of \emph{PL0} were correctly classified as \emph{PL0} and $87\%$ of \emph{PL2} were correctly classified as \emph{PL2}. Respective F1-scores were $83\%/88\%$ (\emph{PL0}/\emph{PL2}). For $\text{PDET}_{012}$, recalls (for 451 aligned words out of 184 utterances) of \emph{PL0}/\emph{PL2} were worse with $79\%$/$62\%$. There were also strong confusions with respect to \emph{PL1} where only $49\%$ of \emph{PL1} were correctly classified as \emph{PL1} but $30\%$ as \emph{PL0} and $21\%$ as \emph{PL2}. %Interestingly, 
\par
For conversation with ID 004M024F, it was also possible to evaluate prominence detection results with respect to the human-annotated labels by keeping only the prominence level information plus word boundary markers in the hypothesis text of lexicon-free (\textbf{Lexfree}) ASR models (cf. ASR experiments in Sec.~\ref{ASRProminence:PDETASR}). More precisely, prominence levels were assigned by majority voting of strings between word boundaries (e.g., the hypothesis "\textbar\phantom{0}d0 i0 e0 \textbar" becomes the string "000" which was assigned as \emph{PL0} but the hypothesis "\textbar\phantom{0}d0 i1 e \textbar" becomes the string "01" which was assigned as an empty string because no clear assignment of a prominence level can be made due to the ambiguity). This results in worse PERs of $65.42\%$ for $\text{ASR}_{02}(\text{PDET}_{02})$ compared to $\text{PDET}_{02}$, partly because not every hypothesis necessarily contains prominence information. This is also reflected in the quality of the alignments for which only $52.17\%$ ($\text{ASR}_{02}(\text{PDET}_{02})$) of the utterances were aligned with respect to word boundaries. Nevertheless, the accuracy of $85.53\%$ of $\text{ASR}_{02}(\text{PDET}_{02})$ demonstrates comparable results to the original prominence detection model $\text{PDET}_{02}$.

\section{Prominence-aware ASR}
\label{ASRProminence:PDETASR}
\subsection{Materials and Methods}
\label{ASRProminence:PDETASR:M&M}

\subsubsection{Data preparation}
\par
\par % prominence annotations
Prominence-aware ASR systems were based on labeled speech data from the entire GRCS component. Pre-processing involved the exclusion of utterances containing laughter, singing, imitations/onomatopoeia, unintelligible word tokens and artefacts which resulted in approx. $\SI{14.4}{\hour}$ (relating to $33734$ utterances) of GRCS data. We standardized typical backchannels (\texttt{mh}, \texttt{hm}, \texttt{mmh}, \texttt{hhm}, \texttt{uh huh}) to \texttt{mhm}, removed punctuation marks and standardized the text to lowercase.

\begin{table*}[t!]
\caption{Concept of character-based prominence-aware ASR training. Generally, each character in the reference text was assigned with a detected prominence level if possible or desired. ASR systems based on $\text{PDET}_{02}$ allow training with a maximum number of two prominence levels (i.e., leading to the systems $\text{ASR}_{0}(\text{PDET}_{02})$, $\text{ASR}_{2}(\text{PDET}_{02})$ and $\text{ASR}_{02}(\text{PDET}_{02})$).}
\centering
%\footnotesize 
\begin{tabular}{c|c|c}
\multicolumn{1}{c|}{\textbf{Type}} & \textbf{Orthography} & \textbf{Reference} \\
\midrule
Baseline & \textbar\phantom{0}die \textbar\phantom{0}waren \textbar\phantom{0}alle \textbar & \textbar\phantom{0}d i e \textbar\phantom{0}w a r e n \textbar\phantom{0}a l l e \textbar \\
$\text{ASR}_{0}(\text{PDET}_{02}$) & \textbar\phantom{0}die \textbar\phantom{0}waren \textbar\phantom{0}alle \textbar & \textbar\phantom{0}d0 i0 e0 \textbar\phantom{0}w0 a0 r0 e0 n0 \textbar\phantom{0}a l l e \textbar \\
$\text{ASR}_{2}(\text{PDET}_{02}$) & \textbar\phantom{0}die \textbar\phantom{0}waren \textbar\phantom{0}alle \textbar & \textbar\phantom{0}d i e \textbar\phantom{0}w a r e n \textbar\phantom{0}a2 l2 l2 e2 \textbar \\
$\text{ASR}_{02}(\text{PDET}_{02}$) & \textbar\phantom{0}die \textbar\phantom{0}waren \textbar\phantom{0}alle \textbar & \textbar\phantom{0}d0 i0 e0 \textbar\phantom{0}w0 a0 r0 e0 n0 \textbar\phantom{0}a2 l2 l2 e2 \textbar \\
\end{tabular}
\label{tab:PDETASROverview}
\end{table*}

\begin{table*}[t!]
\caption{WERs $[\%]$ of two conversations (003M023F/004M024F) for baseline experiments and ASR experiments based on prominence annotations from the prominence detector $\text{PDET}_{02}$.}
\centering
%\footnotesize 
%\setlength{\tabcolsep}{0.5pt}
\scalebox{0.9}{
\begin{tabular}{c|c|c|c}
   & \textbf{Lexfree} & \textbf{Lex} & \textbf{3-gram} \\ 
    \textbf{Type} & \textbf{003M023F}/\textbf{004M024F} & \textbf{003M023F}/\textbf{004M024F}  & \textbf{003M023F}/\textbf{004M024F}  \\  
\midrule
$\text{Baseline}$ & $\textbf{26.04}$ / $\textbf{31.25}$ & $\textbf{21.78}$ / $\textbf{27.52}$ & $18.57$ / $\textbf{23.71}$ \\  \midrule
$\text{ASR}_{0}(\text{PDET}_{02})$ & $26.54$ / $32.32$ & $22.31$ / $28.64$ & $18.58$ / $24.50$ \\
$\text{ASR}_{2}(\text{PDET}_{02})$ & $26.27$ / $32.34$ & $22.24$ / $28.31$ & $18.50$ / $24.32$ \\
$\text{ASR}_{02}(\text{PDET}_{02})$ & $26.66$ / $33.33$ & $23.92$ / $29.84$ & $18.95$ / $25.61$ \\ 
\end{tabular}
}
\label{tab:PDETASRResults}
\end{table*}

\subsubsection{Fine-tuning the prominence-aware ASR}
\par % asr fine-tuning
For all ASR systems, we fine-tuned the pre-trained XLSR model \cite{baevski2020wav2vec,conneau21_interspeech} with a CTC loss \cite{Graves2006}. First, we trained a baseline 
%wav2vec2 
model by mapping the orthography directly to character sequences. Second, we trained prominence-aware ASR systems by including additional information of prominence levels derived from the prominence detectors $\text{PDET}_{02}$ and $\text{PDET}_{012}$. Tab.~\ref{tab:PDETASROverview} shows how the automatic annotations were incorporated into the character-based models by modifying the reference text such that the orthographic reference word sequence also includes character-level prominence information. Thus, for ASR systems based on automatic annotations from $\text{PDET}_{02}$, we trained models which include 1) only prominence level \emph{PL0} ($\text{ASR}_{0}$ with $\approx69$ character tokens\footnote{Note that the number of character tokens can vary with respect to a given training set.}), 2) only prominence level \emph{PL2} ($\text{ASR}_{2}$ with $\approx69$ character tokens\footnotemark[\value{footnote}]), or 3) both prominence levels \emph{PL0}/\emph{PL2} ($\text{ASR}_{02}$ with $\approx102$ character tokens\footnotemark[\value{footnote}]).

For decoding, we used a greedy decoder (\textbf{Lexfree}) and a beam-search decoder with (\textbf{Lex}) and without language model weighting (\textbf{3-gram}). We utilized a consistent lexicon across all models by mapping GRCS words to their corresponding character sequences. Potential prominence levels were only present in the \textbf{Lexfree} outputs, as the beam search decoder was constrained to lexical entries that did not include prominence information. We made this choice because including prominence levels in the lexicon did not improve ASR performance. Consequently, our 
%prominence-aware 
novel ASR system is capable of generating prominence information only 
%when using 
with greedy decoding. The 3-gram LMs were trained with data from each training split with the KenLM toolkit \cite{heafield-2011-kenlm} by using modified Kneser-Ney smoothing and default pruning. We evaluated ASR results on two conversations, namely conversation with ID 003M023F (which was not part of the prominence-annotated subset) and conversation with ID 004M024F (which was also part of the prominence-annotated subset). All ASR results are compared to a wav2vec2 baseline without prominence information ($\approx37$ character tokens\footnotemark[\value{footnote}]).

\subsection{Results for prominence-aware ASR}
Tab.~\ref{tab:PDETASRResults} shows resulting WERs of a baseline and prominence-aware ASR systems for conversations with IDs 003M023F and 004M024F. For the baseline experiments without prominence information, WERs ranged between $18.57\%-26.04\%$ (003M023F) and $23.71\%-31.25\%$ (003M023F). In general, WERs of prominence-aware ASR systems were worse than the baseline systems with absolute maximum deterioration of $2.1\%-2.3\%$ in case of $\text{ASR}_{02}(\text{PDET}_{02})$ and \textbf{Lex}. An exception was the WER of $\text{ASR}_{0}(\text{PDET}{012})$, which was better than the baseline at $18.23\%$, but this improvement occurred only when decoding with a lexicon and LM (003M023F). Worse WERs with detoriations of approx. $1.6\%-2.3\%$ were more likely to occur for systems $\text{ASR}_{02}(\text{PDET}_{02})$ and $\text{ASR}_{02}(\text{PDET}_{012})$ which were based on $\approx 65$ more character tokens in comparison to the baseline systems. Overall, the results indicate that the prominence-aware ASR systems have comparable performance to the baseline systems.

\section{Discussion and conclusion}
\label{ASRProminence:Discussion}

This paper is the first to present a prominence-aware ASR system that can simultaneously transcribe speech and annotate strongly prominent words in spontaneous, conversational speech. With the starting point of a small data set with manually annotated prominence-level labels,  we build a prominence detector to annotate 19h of orthographically annotated conversations. In a subsequent step, we fine-tuned a wav2vec2-based ASR system with speech that contained per word the orthographic annotation and the prominence labels.
%The prominence detection results show similar trends as previously described prominence classification results described in Sec.~\ref{prom_sec:CS_ICPHS}, even though the results are not directly comparable because of differences in the utilized data (cf. Sec.~\ref{prom_sec:CS_ICPHS:limitations}) and the evaluation methods. 
 Our results show that prominence detection was best for a detector that only distinguishes unaccented (PL0) and strongly accented (PL2) words (accuracies of $89.72\% \pm 3.26\%$ for correctly recognized words), indicating that promising detection results can be achieved for both prominence levels. A detector to distinguish PL0, medium-accented (PL1) and PL2 words achieved worse accuracies of $69.45\% \pm 2.11\%$ for correctly recognized words. These findings are in line with what we observed for the inter-annotator agreements for PL1 which had Cohen's kappa of $0.72$ and $0.57$ with respect to \emph{PL1} (cf. Sec.~\ref{ASRProminence:PDET:M&M} and Fig.~\ref{fig:CM_PDET_004M024F}). 
 \par
Heckmann et al. \cite{2014Heckmann} found that despite using different HMM-based alignment strategies for prominence detection, the unweighted accuracies for distinguishing prominent from non-prominent words with prosodic features were approx. $80\%-82\%$, which is in line with our findings. Whereas our prominence detector aligns speech directly to a sequence of prominence levels, the methods in \cite{2014Heckmann} rely on forced alignments that require text transcriptions as input in order to train prominence classifiers. This also implies that their evaluation assumes that all words can be consistently aligned with the human annotations. To conclude, our approach to prominence detection on conversational speech with wav2vec2 works well even without requiring forced alignments to detect phone boundaries. Moreover, our results provide further evidence for fine-tuned speech representation models to capture prosodic information (e.g., \cite{bentum24_interspeech}). We then used this prosodic information for an ASR tasks. Given the unreliable manually created PL1 labels,  we thus only distinguished unaccented from strongly accented words with the prominence-aware ASR system.
\par 
When incorporating the information about (un-)accented words into the wav2vec2-based ASR system, we observed no WER degradation in comparison to the baseline, despite the increased search space, while enabling the transcription of words along with their prominence level. Independent of the decoding strategy (without/with lexicon/LM), the additional prominence information mapped onto the character-level led to consistent results when comparing the WERs to the baseline. However, slightly worse results were achieved for those ASR models where more character tokens were involved. Notably, while the overall prominence error rate (PER) of the prominence-aware ASR system was relatively high at $65.42\%$, our analysis revealed an important finding: in utterances where the ASR-generated word sequence had the correct number of words per utterance, the prominence detection accuracy reached $85.53\%$. This shows that the prominence detection results are highly reliable when the ASR system correctly identifies the number of spoken words, despite the overall higher PER. 
\par 
To conclude, our study demonstrates that prominence detection in conversational speech using wav2vec2 is feasible without relying on forced alignments, as the model effectively extracts prosodic information automatically. When using wav2vec2 for transcribing words and prominence levels simultaneously, the explicit information about prominence levels did not affect ASR performance, while additionally providing labels for prominence levels. To the best of our knowledge, this kind of prominence-enhanced ASR transcript is a novel contribution to the field, with high relevance to both speech science and speech technology. 
Several promising directions for future research emerge from this work. First, our tool could be particularly valuable for linguistic research, especially considering that our approach requires relatively small human-annotated subsets. This could enable efficient prominence annotation for various languages where extensive training data might not be available. Second, our findings could enhance assistive technologies, such as subtitling systems for the deaf and hard of hearing, where prominence-enhanced transcripts could better convey the emphasis of specific words in utterances. Third, this work has implications for prosody-informed dialogue systems, where incorporating prominence information into ASR and NLU components could help automated systems to better understand not just what words were spoken, but also their relative importance for the discourse. This could lead to more nuanced and contextually appropriate responses in human-robot interaction. These applications underscore the broader impact of our contribution to speech science and to applications in speech technology.

\clearpage
\newpage
\section{Acknowledgements}
This research was funded in part by the Austrian Science Fund (FWF) [10.55776/P32700].

\bibliographystyle{IEEEtran}
\bibliography{mybib}

\end{document}